\documentclass[11pt,a4paper]{article}
\usepackage[hyperref]{acl2019}
\usepackage{times}
\usepackage{latexsym}

\usepackage{url}

\usepackage{adjustbox}
\usepackage{graphicx}
\usepackage{amsmath}
\usepackage{amssymb}

\usepackage{color}
\definecolor{amber}{rgb}{1.0, 0.49, 0.0}
\definecolor{applegreen}{rgb}{0.55, 0.71, 0.0}

\usepackage{subcaption}

\newcommand{\beq}{\vspace{0mm}\begin{equation}}
\newcommand{\eeq}{\vspace{0mm}\end{equation}}
\newcommand{\beqs}{\vspace{0mm}\begin{eqnarray}}
\newcommand{\eeqs}{\vspace{0mm}\end{eqnarray}}
\newcommand{\barr}{\begin{array}}
\newcommand{\earr}{\end{array}}

\newcommand{\Fmat}{{\bf F}}

\newcommand{\Mmat}{{\bf M}}

\newcommand{\Umat}{{\bf U}}

\newcommand{\Wmat}{{\bf W}}

\newcommand{\av}{{\boldsymbol a}}

\newcommand{\cv}{{\boldsymbol c}}
\newcommand{\dv}{{\boldsymbol d}}

\newcommand{\hv}{{\boldsymbol h}}

\newcommand{\pv}{{\boldsymbol p}}
\newcommand{\qv}{{\boldsymbol q}}
\newcommand{\rv}{{\boldsymbol r}}
\newcommand{\sv}{{\boldsymbol s}}

\newcommand{\uv}{{\boldsymbol u}}
\newcommand{\vv}{{\boldsymbol v}}

\newcommand{\xv}{{\boldsymbol x}}

\newcommand{\zv}{{\boldsymbol z}}

\newcommand{\alphav}{{\boldsymbol \alpha}}
\newcommand{\betav}{{\boldsymbol \beta}}
\newcommand{\gammav}{{\boldsymbol \gamma}}

\newcommand{\omegav}{{\boldsymbol \omega}}

\newcommand{\R}{\mathbb{R}}

\newcommand{\Acal}{\mathcal{A}}

\aclfinalcopy 

\title{Multi-step Reasoning via Recurrent Dual Attention for Visual Dialog}

\author{Zhe Gan$^{1}$, Yu Cheng$^{1}$, Ahmed El Kholy$^{1}$, Linjie Li$^{1}$, Jingjing Liu$^{1}$, Jianfeng Gao$^{2}$ \\
  $^{1}$Microsoft Dynamics 365 AI Research, \quad$^{2}$Microsoft Research \\
  \small{\texttt{\{zhe.gan, yu.cheng, ahmed.eikholy, lindsey.li, jingjl, jfgao\}@microsoft.com }} }

\date{}

\begin{document}
\maketitle
\begin{abstract}
  This paper presents a new model for visual dialog, Recurrent Dual Attention Network (ReDAN), using multi-step reasoning to answer a series of questions about an image. In each question-answering turn of a dialog, ReDAN infers the answer progressively through multiple reasoning steps. In each step of the reasoning process, the semantic representation of the question is updated based on the image and the previous dialog history, and the recurrently-refined representation is used for further reasoning in the subsequent step. 
  On the VisDial v1.0 dataset, the proposed ReDAN model  
  achieves a new state-of-the-art of 64.47\% NDCG score. 
  Visualization on the reasoning process further demonstrates that ReDAN can locate context-relevant visual and textual clues via iterative refinement, which can lead to the correct answer step-by-step.
\end{abstract}

\section{Introduction}
There has been a recent surge of interest in developing neural network models capable of understanding both visual information and natural language, with applications ranging from image captioning~\cite{fang2015captions,vinyals2015show,xu2015show} to visual question answering (VQA)~\cite{antol2015vqa,fukui2016multimodal,anderson2018bottom}. Unlike VQA, where the model can answer a single question about an image, a visual dialog system~\cite{das2017visual,de2017guesswhat,das2017learning} is designed to answer a series of questions regarding an image, which requires a comprehensive understanding of both the image and previous dialog history.

\begin{figure*}[t]
	\centering
	\begin{subfigure}{.53\textwidth}
		\centering
		\includegraphics[width=1.0\linewidth]{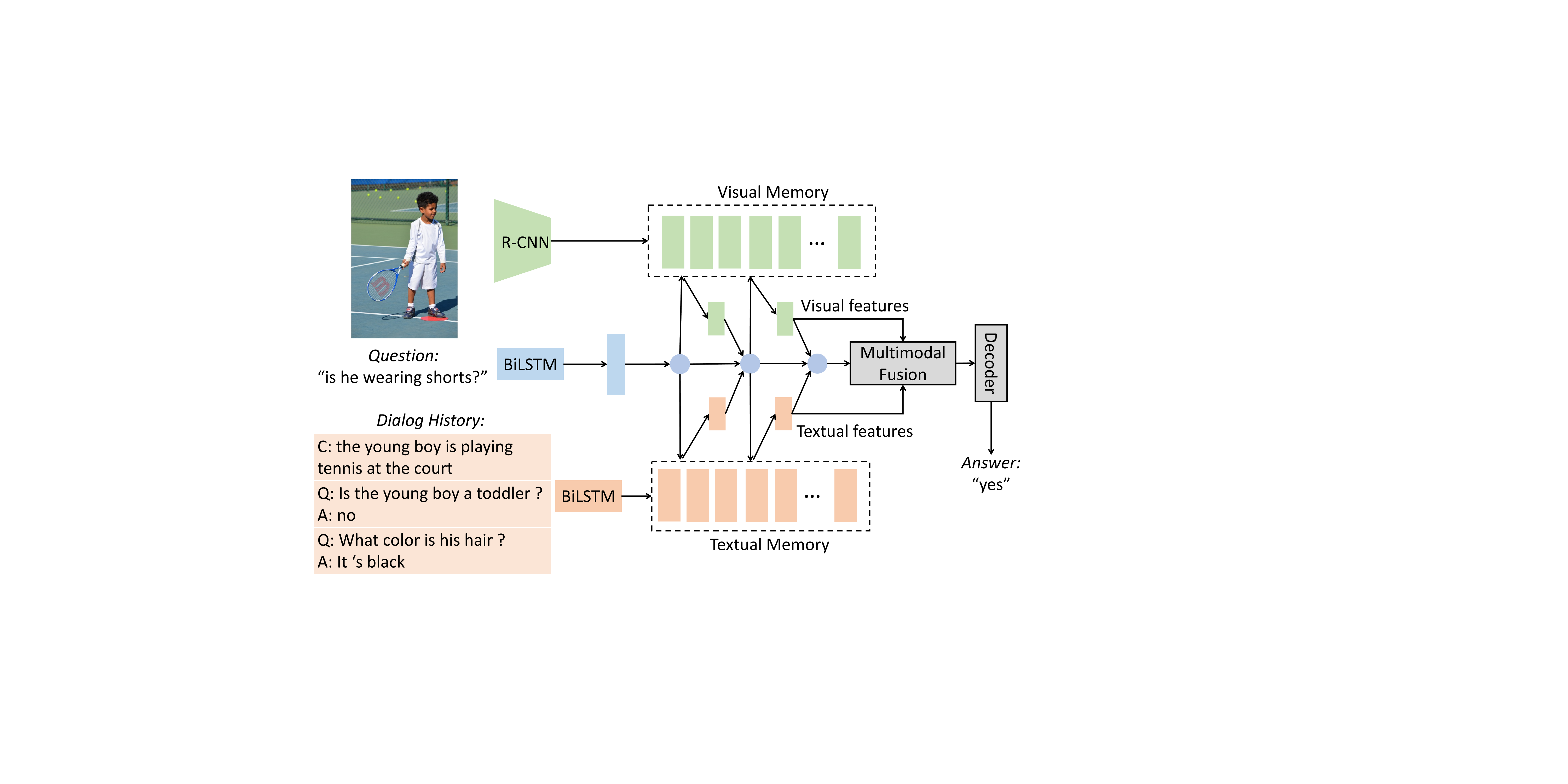}
		\caption{\small{Overview of the proposed ReDAN framework.}}
		\label{fig:overview}
	\end{subfigure}
	\begin{subfigure}{.46\textwidth}
		\centering
		\includegraphics[width=1.0\linewidth]{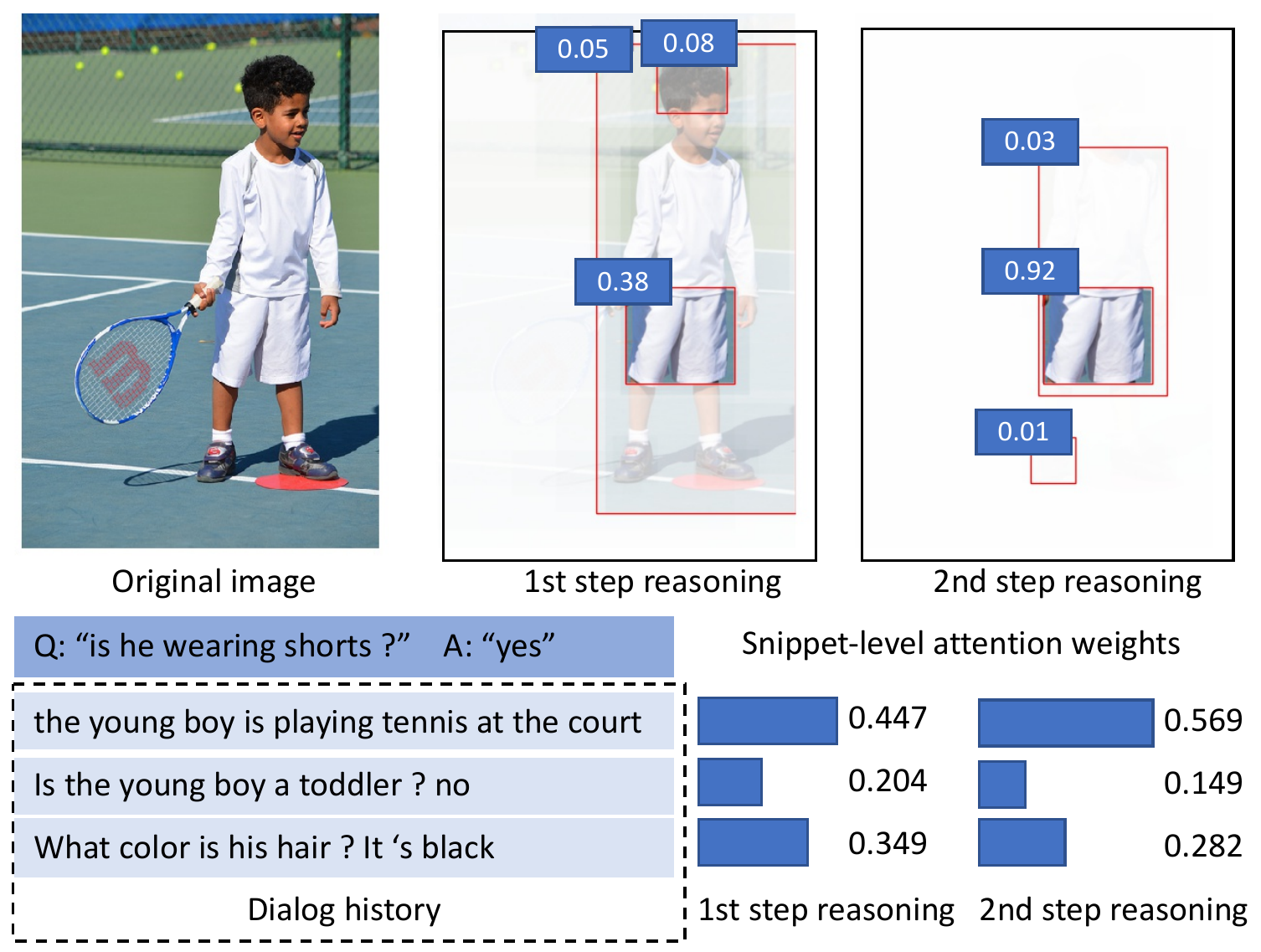}
		\caption{\small{An example of multi-step reasoning in ReDAN.}}
		\label{fig:illustration}
	\end{subfigure}
	\caption{\small{Model architecture and visualization of the learned multi-step reasoning strategies. In the first step, ReDAN first focuses on all relevant objects in the image (\emph{e.g.}, ``\emph{boy}'', ``\emph{shorts}''), and all relevant facts in the dialog history (\emph{e.g.}, ``\emph{young boy}'', ``\emph{playing tennis}'', ``\emph{black hair}''). In the second step, the model narrows down to more context-relevant regions and dialog context (\emph{i.e.}, the attention maps become sharper) which lead to the final correct answer (``\emph{yes}''). The numbers in the bounding boxes and in the histograms are the attention weights of the corresponding objects or dialog history snippets.}} 
	\label{fig:framework}
\end{figure*}

Most previous work on visual dialog rely on attention mechanisms~\cite{bahdanau2014neural,xu2015show} to identify specific regions of the image and dialog-history snippets that are relevant to the question. These attention models measure the relevance between the query and the attended image, as well as the dialog context. To generate an answer, either a discriminative decoder is used for ranking answer candidates, or a generative decoder is trained for synthesizing an answer~\cite{das2017visual,lu2017best}. Though promising results have been reported, these models often fail to provide accurate answers, especially in cases where answers are confined to particular image regions or dialog-history snippets. 

One hypothesis for the cause of failure is the inherent limitation of single-step reasoning approach. 
Intuitively, after taking a first glimpse of the image and the dialog history, readers often revisit specific sub-areas of both image and text to obtain a better understanding of the multimodal context. Inspired by this, we propose a Recurrent Dual Attention Network (ReDAN) that exploits multi-step reasoning for visual dialog. 

Figure~\ref{fig:overview} provides an overview of the model architecture of ReDAN. First, a set of visual and textual memories are created to store image features and dialog context, respectively. In each step, a semantic representation of the question is used to attend to both memories, in order to obtain a question-aware image representation and question-aware dialog representation, both of which subsequently contribute to updating the question representation via a recurrent neural network. Later reasoning steps typically provide a sharper attention distribution than earlier steps, aiming at narrowing down the regions most relevant to the answer. Finally, after several iterations of reasoning steps, the refined question vector and the garnered visual/textual clues are fused to obtain a final multimodal context vector, which is fed to the decoder for answer generation. This  multi-step reasoning process is performed in each turn of the dialog.

Figure~\ref{fig:illustration} provides an illustration of the iterative reasoning process. In the current dialog turn for the question ``\emph{is he wearing shorts?}'', in the initial reasoning step, the system needs to draw knowledge from previous dialog history to know who ``\emph{he}'' refers to (\emph{i.e.}, ``\emph{the young boy}''), as well as interpreting the image to rule out objects irrelevant to the question (\emph{i.e.}, \emph{``net'', ``racket'' and ``court''}). After this, the system conducts a second round of reasoning to pinpoint the image region (\emph{i.e.}, ``\emph{shorts}'', whose attention weight increases from 0.38 to 0.92 from the 1st step to the 2nd step) and the dialog-history snippet (\emph{i.e.}, ``\emph{playing tennis at the court}'', whose attention weight increased from 0.447 to 0.569), which are most indicative of the correct answer (``\emph{yes}'').

The main contributions of this paper are three-fold. ($i$) We propose a ReDAN framework that supports multi-step reasoning for visual dialog. ($ii$) We introduce a simple rank aggregation method to combine the ranking results of discriminative and generative models to further boost the performance. ($iii$) Comprehensive evaluation and visualization analysis demonstrate the effectiveness of our model in inferring answers progressively through iterative reasoning steps.
Our proposed model achieves a new state-of-the-art of 64.47\% NDCG score on the VisDial v1.0 dataset.


\section{Related Work}
\paragraph{Visual Dialog}
The visual dialog task was recently proposed by~\citet{das2017visual} and~\citet{de2017guesswhat}. Specifically,~\citet{das2017visual} released the VisDial dataset, which contains free-form natural language questions and answers.
And~\citet{de2017guesswhat} introduced the GuessWhat$?!$ dataset, where the dialogs provided are more goal-oriented and aimed at object discovery within an image, through a series of yes/no questions between two dialog agents. 

For the VisDial task,  a typical system follows the encoder-decoder framework proposed in~\citet{sutskever2014sequence}. Different encoder models have been explored in previous studies, including late fusion, hierarchical recurrent network, memory network (all three proposed in~\citet{das2017visual}), early answer fusion~\cite{jain2018two}, history-conditional image attention~\cite{lu2017best}, and sequential co-attention~\cite{wu2017you}. The decoder model usually falls into two categories: ($i$) generative decoder to synthesize the answer with a Recurrent Neural Network (RNN)~\cite{das2017visual}; and ($ii$) discriminative decoder to rank answer candidates via a softmax-based cross-entropy loss~\cite{das2017visual} or a ranking-based multi-class N-pair loss~\cite{lu2017best}.

Reinforcement Learning (RL) was used in~\citet{das2017learning,chattopadhyay2017evaluating} to train two agents to play image guessing games.~\citet{lu2017best} proposed a training schema to effectively transfer knowledge from a pre-trained discriminative  model to a generative dialog model. Generative Adversarial Network~\cite{goodfellow2014generative,yu2017seqgan,li2017adversarial} was also used in~\citet{wu2017you} to generate answers indistinguishable from human-generated answers, and a conditional variational autoencoder~\cite{kingma2013auto,sohn2015learning} was developed in~\citet{massiceti2018flipdial} to promote answer diversity. There were also studies investigating visual coreference resolution, either via attention memory implicitly \cite{seo2017visual} or using a more explicit reasoning procedure \cite{kottur2018visual} based on neural module networks~\cite{andreas2016neural}. In addition to answering questions, question sequence generation is also investigated in~\citet{jain2018two,massiceti2018flipdial}.

For the GuessWhat$?!$ task, various methods (such as RL) have been proposed to improve the performance of dialog agents, measured by task completion rate as in goal-oriented dialog system~\cite{strub2017end,shekhar2018ask,strub2018visual,lee2018answerer,zhang2018goal}. Other related work includes image-grounded chitchat~\cite{mostafazadeh2017image}, dialog-based image retrieval~\cite{guo2018dialog}, and text-only conversational question answering~\cite{reddy2018coqa,choi2018quac}. A recent survey on neural approaches to dialog modeling can be found in ~\citet{gaosurvey}.

In this work, we focus on the VisDial task. Different from previous approaches to visual dialog, which all used a single-step reasoning strategy, we propose a novel multi-step reasoning framework that can boost the performance of visual dialog systems by inferring context-relevant information from the image and the dialog history iteratively. 

\begin{figure*}
	\centering
	\includegraphics[width=0.99\linewidth]{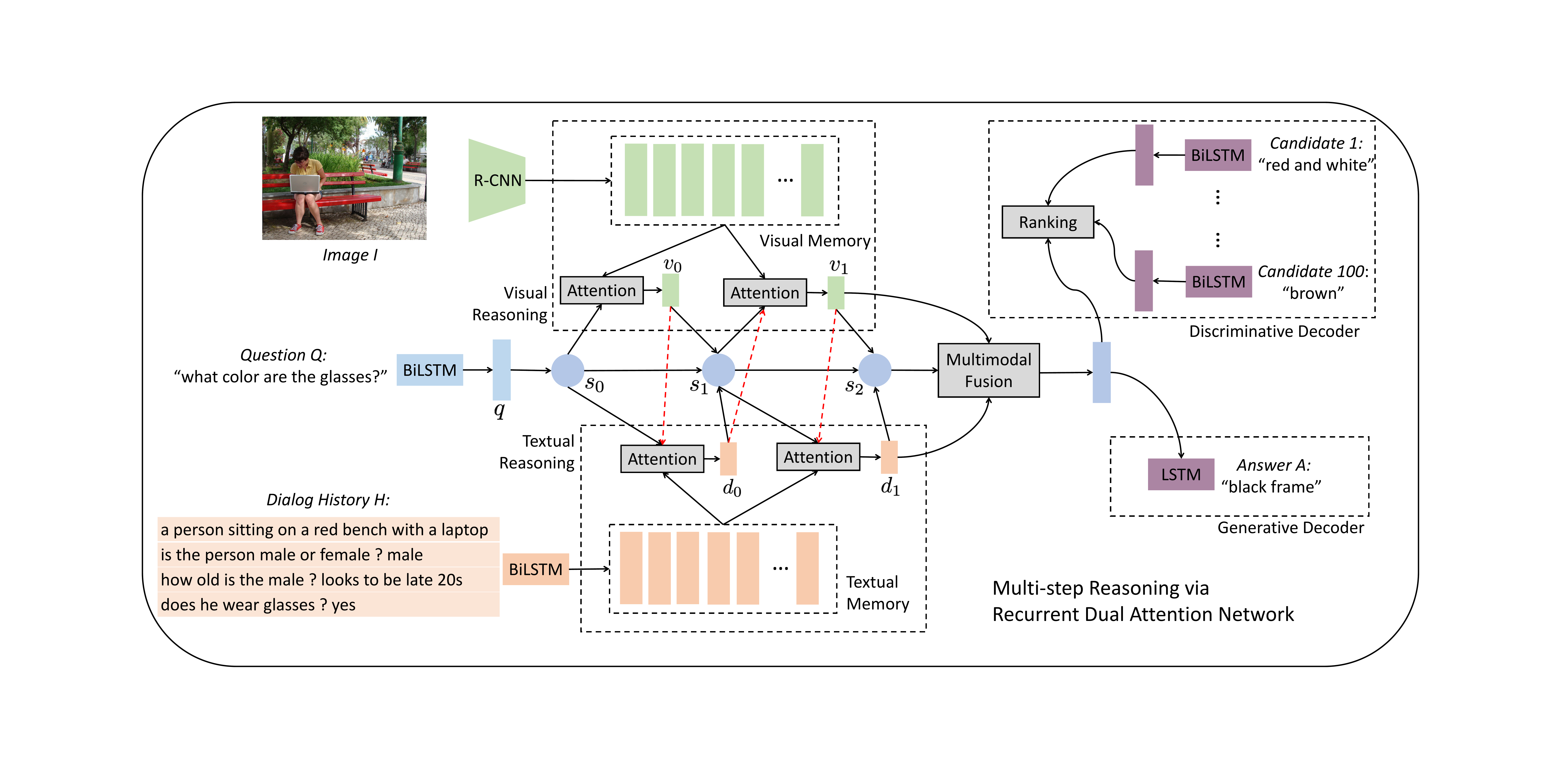}
	\caption{\small{Model Architecture of Recurrent Dual Attention Network for visual dialog. Please see Sec.~\ref{sec:method} for details. }}
	\label{fig:framework}
\end{figure*}

\paragraph{Multi-step Reasoning}
The idea of multi-step reasoning has been explored in many tasks, including image classification~\cite{mnih2014recurrent}, text classification~\cite{yu2017learning}, image generation~\cite{gregor2015draw}, language-based image editing~\cite{chen2017language}, Visual Question Answering (VQA)~\cite{yang2016stacked,nam2016dual,hudson2018compositional}, and Machine Reading Comprehension (MRC)~\cite{cui2016attention,dhingra2016gated,hill2015goldilocks,sordoni2016iterative,shen2017reasonet,liu2017stochastic}.

Specifically,~\citet{mnih2014recurrent} introduced an RNN for image classification, by selecting a sequence of regions adaptively and only processing the selected regions.~\citet{yu2017learning} used an RNN for text classification, by learning to skip irrelevant information when reading the text input. A recurrent variational autoencoder termed DRAW was proposed in~\citet{gregor2015draw} for multi-step image generation. A recurrent attentive model for image editing was also proposed in~\citet{chen2017language} to fuse image and language features via multiple steps. 

For VQA, Stacked Attention Network (SAN)~\cite{yang2016stacked} was proposed to attend the question to relevant image regions  via multiple attention layers. For MRC, ReasoNet~\cite{shen2017reasonet} was developed to perform multi-step reasoning to infer the answer span based on a given passage and a question, where the number of steps can be dynamically determined via a termination gate. 

Different from SAN for VQA \cite{yang2016stacked} and ReasoNet for MRC \cite{shen2017reasonet}, which reason over a single type of input (either image or text), our proposed ReDAN model incorporates multimodal context that encodes both visual information and textual dialog. This multimodal reasoning approach presents a mutual enhancement between image and text for a better understanding of both: on the one hand, the attended image regions can provide additional information for better dialog interpretation; on the other hand, the attended history snippets can be used for better image understanding (see the dotted red lines in Figure~\ref{fig:framework}).

\paragraph{Concurrent Work} We also include some concurrent work for visual dialog that has not been discussed above, including image-question-answer synergistic network~\cite{guo2019image}, recursive visual attention~\cite{niu2018recursive}, factor graph attention~\cite{schwartz2019factor}, dual attention network~\cite{kang2019dual}, graph neural network~\cite{zheng2019reasoning}, history-advantage sequence training~\cite{yang2019making}, and weighted likelihood estimation~\cite{zhang2019generative}. 

\section{Recurrent Dual Attention Network} \label{sec:method}
The visual dialog task~\cite{das2017visual} is formulated as follows: given a question $Q_\ell$ grounded in an image $I$, and previous dialog history (including the image caption $C$) $H_\ell = \{C,(Q_1, A_1), \cdots, (Q_{\ell-1},A_{\ell-1})\}$ ($\ell$ is the current dialog turn) as additional context, the goal is to generate an answer by ranking a list of $N$ candidate answers $\Acal_\ell = \{A_\ell^{(1)},\ldots, A_\ell^{(N)}\}$. 

Figure~\ref{fig:framework} provides an overview of the Recurrent Dual Attention Network (ReDAN). Specifically, ReDAN consists of three components: ($i$) \emph{Memory Generation Module} (Sec.~\ref{sec:memory}), which generates a set of visual and textual memories to provide grounding for reasoning; ($ii$) \emph{Multi-step Reasoning Module} (Sec.~\ref{sec:reasoning}), where recurrent dual attention is applied to jointly encode question, image and dialog history into a multimodal context vector for decoding; and ($iii$) \emph{Answer Decoding Module} (Sec.~\ref{sec:decoding}), which derives the final answer for each question based on the multimodal context vector. The following sub-sections describe the details of these components. 

\subsection{Memory Generation Module} \label{sec:memory}
In this module, the image $I$ and the dialog history $H_\ell$ are transformed into a set of memory vectors (visual and textual). 

\paragraph{Visual Memory}
We use a pre-trained Faster R-CNN~\cite{ren2015faster,anderson2018bottom} to extract image features, in order to enable attention on both object-level and salient region-level, each associated with a feature vector. Compared to image features extracted from VGG-Net~\cite{simonyan2014very} and ResNet~\cite{he2016deep}, this type of features from Faster R-CNN has achieved state-of-the-art performance in both image captioning and VQA~\cite{anderson2018bottom,teney2017tips} tasks. Specifically, the image features $\Fmat_I$ for a raw image $I$ are represented by: 
\begin{align}
\Fmat_I = \mbox{R-CNN} (I) \in \R^{n_f\times M} \,,
\end{align}
where $M=36$ is the number of detected objects in an image\footnote{We have also tried using an adaptive number of detected objects for an image. Results are very similar to the results with $M=36$.}, and $n_f=2048$ is the dimension of the feature vector. A single-layer perceptron is used to transform each feature into a new vector that has the same dimension as the query vector (described in Sec.~\ref{sec:reasoning}):
\begin{align}
\Mmat_v = \tanh (\Wmat_I \Fmat_I) \in \R^{n_h \times M} \,,
\end{align}
where $\Wmat_I \in \R^{n_h \times n_f}$. All the bias terms in this paper are omitted for simplicity. $\Mmat_v$ is the visual memory, and its $m$-th column corresponds to the visual feature vector for the region of the object indexed by $m$.

\paragraph{Textual Memory}
In the $\ell$-th dialogue turn, the dialog history $H_\ell$ consists of the caption $C$ and $\ell-1$ rounds of QA pairs $(Q_j, A_j)$ ($j=1,\ldots,\ell-1$). For each dialog-history snippet $j$ (the caption is considered as the first one with $j=0$), it is first represented as a matrix $\Mmat_h^{(j)} = [\hv_0^{(j)}, \ldots, \hv_{K-1}^{(j)}] \in \R^{n_h \times K}$ via a bidirectional Long Short-Term Memory (BiLSTM) network~\cite{hochreiter1997long}, where $K$ is the maximum length of the dialog-history snippet. Then, a self-attention mechanism is applied to learn the attention weight of every word in the snippet, identifying the key words and ruling out irrelevant information. Specifically, 
\begin{align}
\omegav_j &= \mbox{softmax}(\pv_{\omega}^T\cdot\tanh(\Wmat_h \Mmat_h^{(j)}) )\,, \nonumber \\ 
\uv_j &= \omegav_j \cdot (\Mmat_h^{(j)})^T\,,
\end{align}
where $\omegav_j \in \R^{1\times K}, \pv_{\omega} \in \R^{n_h\times 1}$, $\Wmat_h \in \R^{n_h\times n_h}$, and $\uv_j \in \R^{1\times n_h}$. After applying the same BiLSTM to each dialog-history snippet, the textual memory is then represented as $\Mmat_d = [\uv_0^T,\ldots,\uv_{\ell-1}^T] \in \R^{n_h\times \ell}$.  


\subsection{Multi-step Reasoning Module} \label{sec:reasoning}
The multi-step reasoning framework is implemented via an RNN, where the hidden state $\sv_t$ represents the current representation of the question, and acts as a query to retrieve visual and textual memories. The initial state $\sv_0$ is a self-attended question vector $\qv$. Let $\vv_t$ and $\dv_t$ denote the attended image representation and dialog-history representation in the $t$-th step, respectively. 
A one-step reasoning pathway can be illustrated as $\sv_t \rightarrow \vv_t \rightarrow \dv_t \rightarrow \sv_{t+1}$, which is performed $T$ times. Details are described below. 

\paragraph{Self-attended Question}
Similar to textual memory construction, a question $Q$ (the subscript $\ell$ for $Q_\ell$ is omitted to reduce confusion) is first represented as a matrix $\Mmat_q = [\qv_0, \ldots, \qv_{K^\prime-1}] \in \R^{n_h \times K^\prime}$ via a BiLSTM, where $K^\prime$ is the maximum length of the question. Then, self attention is applied, 
\begin{align*}
\alphav = \mbox{softmax}(\pv_{\alpha}^T\cdot\tanh(\Wmat_q \Mmat_q) )\,, \,\, \qv = \alphav \Mmat_q^T\,,
\end{align*}
where $\alphav \in \R^{1\times K^\prime}, \pv_{\alpha} \in \R^{n_h\times 1}$, and $\Wmat_q \in \R^{n_h\times n_h}$. $\qv \in \R^{1\times n_h}$ then serves as the initial hidden state of the RNN, \emph{i.e.}, $\sv_0 = \qv$.

The reasoning pathway $\sv_t \rightarrow \vv_t \rightarrow \dv_t \rightarrow \sv_{t+1}$ includes the following steps: ($i$) $(\sv_t, \dv_{t-1}) \rightarrow \vv_t$; ($ii$) $(\sv_t,\vv_t) \rightarrow \dv_t$; and ($iii$) $(\vv_t,\dv_t) \rightarrow \sv_{t+1}$.

\paragraph{Query and History Attending to Image}
Given $\sv_t$ and the previous attended dialog history representation $\dv_{t-1}\in \R^{1\times n_h}$, we update $\vv_t$ as follows:
\begin{adjustbox}{minipage=1.08\linewidth,scale=0.84}
\begin{align}
\betav &= \mbox{softmax}(\pv_{\beta}^T\cdot\tanh(\Wmat_v \Mmat_v + \Wmat_s \sv_t^T + \Wmat_d \dv_{t-1}^T) )\,,\nonumber \\
\vv_t &= \betav \cdot \Mmat_v^T\,,
\end{align}
\end{adjustbox}
where $\betav \in \R^{1\times M}, \pv_{\beta} \in \R^{n_h\times 1}, \Wmat_v \in \R^{n_h\times n_h}, \Wmat_s \in \R^{n_h\times n_h}$ and $\Wmat_d \in \R^{n_h\times n_h}$. The updated $\vv_t$, together with $\sv_t$, is used to attend to the dialog history. 

\paragraph{Query and Image Attending to History}
Given $\sv_t \in \R^{1\times n_h}$ and the attended image representation $\vv_{t} \in \R^{1\times n_h}$, we update $\dv_t$ as follows:

\begin{adjustbox}{minipage=1.08\linewidth,scale=0.88}
\begin{align}
\gammav &= \mbox{softmax}(\pv_{\gamma}^T\cdot\tanh(\Wmat^{'}_d \Mmat_d + \Wmat^{'}_s \sv_t^T + \Wmat^{'}_v \vv_t^T) )\,,\nonumber \\
\dv_t &= \gammav \cdot \Mmat_d^T\,,
\end{align}
\end{adjustbox}
where $\gammav \in \R^{1\times \ell}, \pv_{\gamma} \in \R^{n_h\times 1}, \Wmat^{'}_v \in \R^{n_h\times n_h}, \Wmat^{'}_s \in \R^{n_h\times n_h}$ and $\Wmat^{'}_d \in \R^{n_h\times n_h}$. The updated $\dv_t$ is fused with $\vv_t$ and then used to update the RNN query state. 

\paragraph{Multimodal Fusion}
Given the query vector $\sv_t$, we have thus far obtained the updated image representation $\vv_t$ and the dialog-history representation $\dv_t$. Now, we use Multimodal Factorized Bilinear pooling (MFB)~\cite{yu2017multi} to fuse $\vv_t$ and $\dv_t$ together. Specifically, 
\begin{align}
\zv_t &= \mbox{SumPooling} (\Umat_v\vv_t^T \circ \Umat_d \dv_t^T, k)\,, \label{eqn:sumpool} \\
\zv_t &= \mbox{sign}(\zv_t)|\zv_t|^{0.5}, \,\, \zv_t = \zv_t^T / ||\zv_t||\,, \label{eqn:power_norm}
\end{align}
where $\Umat_v \in \R^{n_h k \times n_h}, \Umat_d \in \R^{n_h k \times n_h}$. The function $\mbox{SumPooling}(\xv,k)$ in (\ref{eqn:sumpool}) means using a one-dimensional non-overlapped window with the size $k$ to perform sum pooling over $\xv$. (\ref{eqn:power_norm}) performs power normalization and $\ell_2$ normalization. The whole process is denoted in short as:
\begin{align}
\zv_t = \mbox{MFB} (\vv_t, \dv_t) \in \R^{1\times n_h} \,.
\end{align}
There are also other methods for multimodal fusion, such as MCB~\cite{fukui2016multimodal} and MLB~\cite{kim2016hadamard}. We use MFB in this paper due to its superior performance in VQA. 

\paragraph{Image and History Updating RNN State}
The initial state $\sv_0$ is set to $\qv$, which represents the initial understanding of the question. The question representation is then updated based on the current dialogue history and the image, via an RNN with Gated Recurrent Unit (GRU)~\cite{cho2014learning}: 
\begin{align}
\sv_{t+1} = \mbox{GRU}(\sv_t, \zv_t)\,.
\end{align}
This process forms a cycle completing one reasoning step. 
After performing $T$ steps of reasoning, multimodal fusion is then used to obtain the final context vector:
\begin{align*}
\cv = [\mbox{MFB} (\sv_T, \vv_T), \mbox{MFB} (\sv_T, \dv_T), \mbox{MFB} (\vv_T, \dv_T)] \,.
\end{align*}
%

\subsection{Answer Decoding Module} \label{sec:decoding}
\paragraph{Discriminative Decoder}
The context vector $\cv$ is used to rank answers from a pool of candidates $\Acal$ (the subscript $\ell$ for $\Acal_\ell$ is omitted). Similar to how we obtain the self-attended question vector in Sec.~\ref{sec:reasoning}, a BiLSTM, together with the self-attention mechanism, is used to obtain a vector representation for each candidate $A_j \in \Acal$, resulting in $\av_j \in \R^{1\times n_h}$, for $j=1, \ldots, N$. 
Based on this, a probability vector $\pv$ is computed as $\pv = \mbox{softmax}(\sv)$, where $\sv \in \R^{N}$, and $\sv[j]=\cv \av_j^T$. During training, ReDAN is optimized by minimizing the cross-entropy loss\footnote{We have also tried the N-pair ranking loss used in~\citet{lu2017best}. Results are very similar to each other.} between the one-hot-encoded ground-truth label vector and the probability distribution $\pv$.
%
During evaluation, the answer candidates are simply ranked based on the probability vector $\pv$. 

\paragraph{Generative Decoder}
Besides the discriminative decoder, following~\citet{das2017visual}, we also consider a generative decoder, where another LSTM is used to decode the context vector into an answer. During training, we maximize the log-likelihood of the ground-truth answers. During evaluation, we use the log-likelihood scores to rank answer candidates.

\paragraph{Rank Aggregation}
Empirically, we found that combining the ranking results of discriminative and generative decoders boosts the performance a lot. 
Two different rank aggregation methods are explored here: ($i$) average over ranks; and ($ii$) average over reciprocal ranks. 
Specifically, in a dialog session, assuming $\rv_1, \ldots, \rv_K$ represents the ranking results obtained from $K$ trained models (either discriminative, or generative). In the first method, the average ranks $\frac{1}{K}\sum_{k=1}^K \rv_k$ are used to re-rank the candidates. In the second one, we use the average of the reciprocal ranks of each individual model $\frac{1}{K}\sum_{k=1}^K 1/\rv_k$ for re-ranking.

\section{Experiments} \label{sec:exps}
In this section, we explain in details our experiments on the VisDial dataset. We compare our ReDAN model with state-of-the-art baselines, and conduct detailed analysis to validate the effectiveness of our proposed model. 

\begin{table*}
	\small
	\begin{center}
		\begin{tabular}{|l|cccccc|}
			\hline
			Model  & NDCG & MRR & R@1 & R@5 & R@10 & Mean\\
			\hline\hline
			MN-D ~\cite{das2017visual}  & 55.13 & 60.42 & 46.09 & 78.14 & 88.05 & 4.63 \\
			HCIAE-D ~\cite{lu2017best}  & 57.65 & 62.96 & 48.94 & 80.50 & 89.66 & 4.24 \\
			CoAtt-D ~\cite{wu2017you}  & 57.72 & 62.91 & 48.86 & 80.41 & 89.83 & 4.21 \\
			\hline \hline
			ReDAN-D  ($T$=1) & 58.49 & 63.35 & 49.47 & 80.72 & 90.05 & 4.19 \\
			ReDAN-D  ($T$=2)  & 59.26 & 63.46 & 49.61 & 80.75 & 89.96 & 4.15 \\
			ReDAN-D  ($T$=3) & 59.32 & 64.21 & 50.60 & 81.39 & 90.26 & 4.05 \\
			\hline \hline
			Ensemble of 4  & \textbf{60.53} & \textbf{65.30}  & \textbf{51.67}  & \textbf{82.40}  & \textbf{91.09} & \textbf{3.82}  \\
			\hline
		\end{tabular}
	\end{center}
	\vspace{-3mm}
	\caption{\small{Comparison of ReDAN with a discriminative decoder to state-of-the-art methods on VisDial v1.0 validation set. Higher score is better for NDCG, MRR and Recall@$k$, while lower score is better for mean rank.
	All these baselines are re-implemented with bottom-up features and incorporated with GloVe vectors for fair comparison.} \label{tab:results_val}}
\end{table*}

\begin{table*}
	\small
	\begin{center}
		\begin{tabular}{|l|cccccc|}
			\hline
			Model  & NDCG & MRR & R@1 & R@5 & R@10 & Mean\\
			\hline\hline
			MN-G~\cite{das2017visual} & 56.99  & 47.83 & 38.01 & 57.49 & 64.08 & 18.76 \\
			HCIAE-G~\cite{lu2017best} & 59.70  & 49.07 & 39.72 & 58.23 & 64.73 & 18.43 \\
			CoAtt-G~\cite{wu2017you}  & 59.24  & 49.64 & 40.09 & 59.37 & 65.92 & 17.86 \\
			\hline \hline
			ReDAN-G ($T$=1)  & 59.41 & 49.60 & 39.95 & 59.32 & 65.97 & 17.79 \\
			ReDAN-G ($T$=2) & 60.11 & 49.96 & 40.36 & 59.72 & 66.57 & 17.53 \\
			ReDAN-G ($T$=3) &  60.47 & 50.02 & 40.27 & 59.93 & 66.78 & 17.40 \\
			\hline \hline
			Ensemble of 4 & \textbf{61.43} & \textbf{50.41} & \textbf{40.85}  & \textbf{60.08}  & \textbf{67.17} & \textbf{17.38} \\
			\hline
		\end{tabular}
	\end{center}
	\vspace{-3mm}
	\caption{\small{Comparison of ReDAN with a generative decoder to state-of-the-art generative methods on VisDial val v1.0. All the baseline models are re-implemented with bottom-up features and incorporated with GloVe vectors for fair comparison.}  \label{tab:results_val_gene}}
\end{table*}

\subsection{Experimental Setup}
\paragraph{Dataset} We evaluate our proposed approach on the recently released VisDial v1.0 dataset\footnote{As suggested in \url{https://visualdialog.org/data}, results should be reported on v1.0, instead of v0.9.}. Specifically, the training and validation splits from v0.9 are combined together to form the new training data in v1.0, which contains dialogs on $123,287$ images from COCO dataset~\cite{lin2014microsoft}. Each dialog is equipped with 10 turns, resulting in a total of 1.2M question-answer pairs.
An additional $10,064$ COCO-like images are further collected from Flickr, of which $2,064$ images are used as the validation set (val v1.0), and the rest 8K are used as the test set (test-std v1.0), hosted on an evaluation server\footnote{\url{https://evalai.cloudcv.org/web/challenges/challenge-page/161/overview}} (the ground-truth answers for this split are not publicly available). 
Each image in the val v1.0 split is associated with a 10-turn dialog, while a dialog with a flexible number of turns is provided for each image in test-std v1.0. Each question-answer pair in the VisDial dataset is accompanied by a list of 100 answer candidates, and the goal is to find the correct answer among all the candidates. 

\paragraph{Preprocessing}
We truncate captions/questions/ answers that are longer than 40/20/20 words, respectively. And we build a vocabulary of words that occur at least 5 times in train v1.0, resulting in $11,319$ words in the vocabulary. For word embeddings, we use pre-trained GloVe vectors~\cite{pennington2014glove} for all the captions, questions and answers,
concatenated with the
learned word embedding from the BiLSTM encoders to further boost the performance.
For image representation, we use bottom-up-attention features~\cite{anderson2018bottom} extracted from Faster R-CNN~\cite{ren2015faster} pre-trained on Visual Genome~\cite{krishna2017visual}. A set of 36 features is created for each image. Each feature is a 2048-dimentional vector. 

\paragraph{Evaluation}
Following~\citet{das2017visual}, we use a set of ranking metrics (Recall@$k$ for $k=\{1,5,10\}$, mean rank, and mean reciprocal rank (MRR)), to measure the performance of retrieving the  ground-truth answer from a pool of 100 candidates. Normalized Discounted Cumulative Gain (NDCG) score is also used for evaluation in the visual dialog challenge 2018 and 2019, based on which challenge winners are picked.
Since this requires dense human annotations, the calculation of NDCG is only available on val v1.0, test-std v1.0, and a small subset of 2000 images from train v1.0.

\paragraph{Training details}
All three BiLSTMs used in the model are single-layer with 512 hidden units. The number of factors used in MFB is set to 5, and we use mini-batches of size 100. The maximum number of epochs is set to 20. No dataset-specific tuning or regularization is conducted except dropout~\cite{srivastava2014dropout} and early stopping on validation sets. The dropout ratio is 0.2. The Adam algorithm~\cite{kingma2014adam} with learning rate $4\times 10^{-4}$ is used for optimization. The learning rate is halved every 10 epochs. 

\begin{figure*}[t!]
	\centering
	\includegraphics[width=1.0\linewidth]{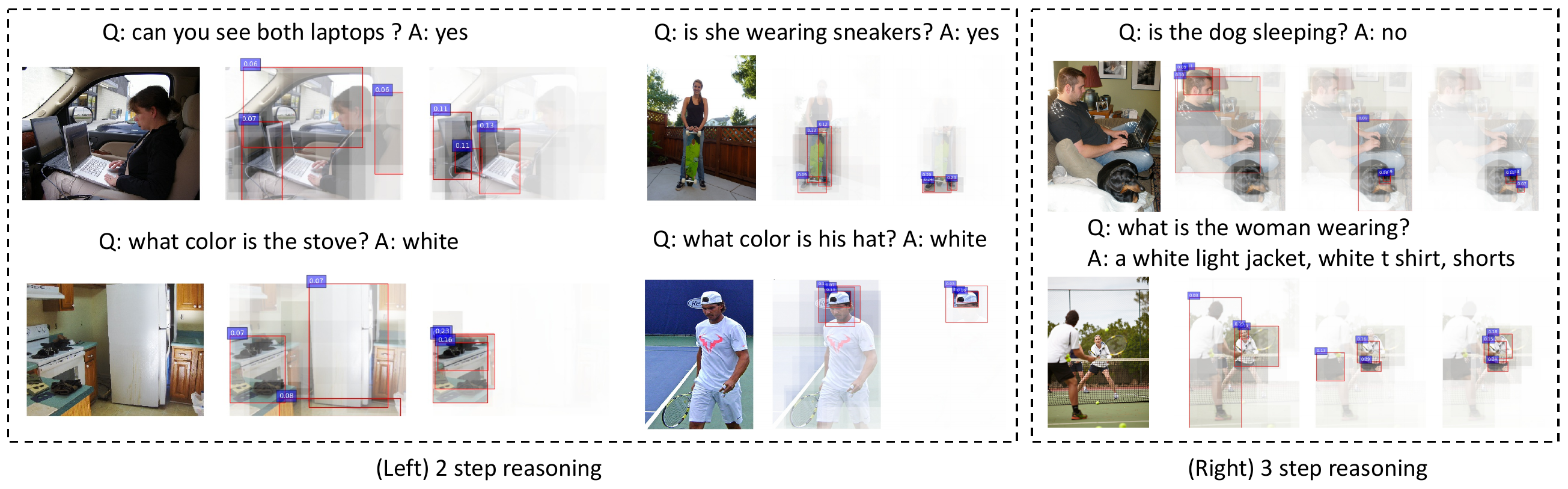}
	\caption{\small{Visualization of learned attention maps in multiple reasoning steps.}}
	\label{fig:example_visual_attention}
\end{figure*}

\subsection{Quantitative Results}
\paragraph{Baselines}
We compare our proposed approach with state-of-the-art models, including Memory Network (MN)~\cite{das2017visual}, History-Conditioned Image Attentive Encoder (HCIAE)~\cite{lu2017best} and Sequential Co-Attention model (CoAtt)~\cite{wu2017you}.
In their original papers, all these models used VGG-Net~\cite{simonyan2014very} for image feature extraction, and reported results on VisDial v0.9. Since bottom-up-attention features have proven to achieve consistently better performance than VGG-Net in other tasks, we re-implemented all these models with bottom-up-attention features, and used the same cross-entropy loss for training. Further, unidirectional LSTMs are used in these previous baselines, which are replaced by bidirectional LSTMs with self-attention mechanisms for fair comparison. All the baselines are also further incorporated with pre-trained GloVe vectors. 
We choose the best three models on VisDial v0.9 as the baselines:
\begin{itemize}
	\item \textbf{MN}~\cite{das2017visual}: ($i$) mean pooling is performed over the bottom-up-attention features for image representation; ($ii$) image and question attend to the dialog history.  
	\item \textbf{HCIAE}~\cite{lu2017best}: ($i$) question attends to dialog history; ($ii$) then, question and the attended history attend to the image.
	\item \textbf{CoAtt}~\cite{wu2017you}: ($i$) question attends to the image; ($ii$) question and image attend to the history; ($iii$) image and history attend to the question; ($iv$) question and history attend to the image again.
\end{itemize}


\begin{table*}[t!]
	\small
	\begin{center}
		\begin{tabular}{|cc|cccccc|}
			\hline
			Model & Ens. Method & NDCG & MRR & R@1 & R@5 & R@10 & Mean \\
			\hline\hline
			4 Dis. & Average & 60.53 & 65.30  & 51.67 & 82.40 & 91.09 & 3.82  \\
			4 Gen. & Average & 61.43 & 50.41  & 40.85 & 60.08 & 67.17 & 17.38  \\
			\hline\hline
			1 Dis. + 1 Gen. & Average & 63.85 & 53.53 & 42.16 & 65.43 & 74.36 & 9.00 \\
			1 Dis. + 1 Gen. & Reciprocal & 63.18 & 59.03 & 42.33 & 78.71 & 88.13 & 4.88 \\
			\hline\hline
			4 Dis. + 4 Gen. & Average & 65.13 & 54.19 & 42.92 & 66.25 & 74.88 & 8.74 \\
			4 Dis. + 4 Gen. & Reciprocal & 64.75 & 61.33 &  45.52 & 80.67 & 89.55 & 4.41 \\
			\hline\hline
			ReDAN+ (Diverse Ens.) & Average & \textbf{67.12} & 56.77 & 44.65 & 69.47 & 79.90 & 5.96 \\
			\hline
		\end{tabular}
	\end{center}
	\vspace{-3mm}
	\caption{\small{Results of different rank aggregation methods. Dis. and Gen. is short for discriminative and generative model, respectively.} \label{tab:ens}}
\end{table*}

\paragraph{Results on VisDial val v1.0}
Experimental results on val v1.0 are shown in Table~\ref{tab:results_val}. ``-D'' denotes that a discriminative decoder is used. With only one reasoning step, our ReDAN model already achieves better performance than CoAtt, which is the previous best-performing model. Using two or three reasoning steps further increases the performance. Further increasing the number of reasoning steps does not help, thus results are not shown.
%
We also report results on an ensemble of 4 ReDAN-D models. 
Significant improvement was observed, boosting NDCG from 59.32 to 60.53, and MRR from 64.21 to 65.30. 

%
In addition to discriminative decoders, we also evaluate our model with a generative decoder. Results are summarized in Table~\ref{tab:results_val_gene}. Similar to Table~\ref{tab:results_val}, ReDAN-G with $T$=3 also achieves the best performance. It is intuitive to observe that ReDAN-D achieves much better results than ReDAN-G on MRR, R@$k$ and Mean Rank, since ReDAN-D is a discriminative model, and utilizes much more information than ReDAN-G. For example, ReDAN-D uses both positive and negative answer candidates for ranking/classification, while ReDAN-G only uses positive answer candidates for generation. However, 
interestingly, ReDAN-G achieves better NDCG scores than ReDAN-D (61.43 vs 60.53). 
We provide some detailed analysis in the question-type analysis section below. 
\begin{table*}
	\small
	\begin{center}
		\begin{tabular}{|l|cccccc|}
			\hline
			Model & NDCG & MRR & R@1 & R@5 & R@10 & Mean\\
			\hline\hline
			ReDAN+ (Diverse Ens.) & \textbf{64.47} & 53.73 & 42.45 & 64.68 & 75.68 & 6.63 \\
			ReDAN (1 Dis. + 1 Gen.) & 61.86 & 53.13 & 41.38 & 66.07 & 74.50 & 8.91 \\
			\hline\hline
			DAN~\cite{kang2019dual} & 59.36 & 64.92 & 51.28 & 81.60 & 90.88 & 3.92 \\
			NMN~\cite{kottur2018visual} & 58.10 & 58.80 & 44.15 & 76.88 & 86.88 & 4.81 \\
			Sync~\cite{guo2019image} & 57.88 & 63.42 & 49.30 & 80.77 & 90.68 & 3.97 \\
			HACAN~\cite{yang2019making} & 57.17 & 64.22 & 50.88 & 80.63 & 89.45 & 4.20 \\
			FGA$^\dag$ & 57.13 & 69.25 & 55.65 & 86.73 & 94.05 & 3.14 \\
			USTC-YTH$^\ddag$ & 56.47 & 61.44 & 47.65 & 78.13 & 87.88 & 4.65 \\
			RvA~\cite{niu2018recursive} & 55.59 & 63.03 & 49.03 & 80.40 & 89.83 & 4.18 \\
			MS ConvAI$^\ddag$ & 55.35 & 63.27 & 49.53 & 80.40 & 89.60 & 4.15 \\
			CorefNMN~\cite{kottur2018visual} & 54.70 & 61.50 & 47.55 & 78.10 & 88.80 & 4.40 \\
			FGA~\cite{schwartz2019factor} & 54.46 & 67.25 & 53.40 & 85.28 & 92.70 & 3.54 \\
			GNN~\cite{zheng2019reasoning} & 52.82 & 61.37 & 47.33 & 77.98 & 87.83 & 4.57 \\
			LF-Att w/ bottom-up$^\dag$ & 51.63 & 60.41 & 46.18 & 77.80 & 87.30 & 4.75 \\
			LF-Att$^\ddag$ & 49.76 & 57.07 & 42.08 & 74.83 & 85.05 & 5.41 \\
			MN-Att$^\ddag$ & 49.58 & 56.90 & 42.43 & 74.00 & 84.35 & 5.59 \\
			MN$^\ddag$  & 47.50 & 55.49 & 40.98 & 72.30 & 83.30 & 5.92 \\
			HRE$^\ddag$  & 45.46 & 54.16 & 39.93 & 70.45 & 81.50 & 6.41 \\
			LF$^\ddag$  & 45.31 & 55.42 & 40.95 & 72.45 & 82.83 & 5.95 \\
			\hline
		\end{tabular}
	\end{center}
	\vspace{-3mm}
	\caption{\small{Comparison of ReDAN to state-of-the-art visual dialog models on the blind test-std v1.0 set, as reported by the test server. ($\dag$) taken from \url{https://evalai.cloudcv.org/web/challenges/challenge-page/161/leaderboard/483}. ($\ddag$) taken from \url{https://evalai.cloudcv.org/web/challenges/challenge-page/103/leaderboard/298}. \label{tab:result_test}}}
\end{table*}

\begin{table*}[t!]
	\small
	\begin{center}
		\begin{tabular}{|c|c|cccc|}
			\hline
			Question Type & All & Yes/no & Number & Color & Others \\
			\hline\hline
			Percentage & 100\% & 75\% & 3\%  & 11\% & 11\%  \\
			\hline\hline
			Dis. & 59.32 & 60.89 & 44.47 & 58.13 & 52.68 \\
			Gen. & 60.42 & 63.49 & 41.09 & 52.16 & 51.45 \\
			4 Dis. + 4 Gen. & 65.13 & 68.04 & 46.61 & 57.49 & 57.50 \\
			ReDAN+ & 67.12 & 69.49 & 50.10 & 62.70 & 58.50 \\
			\hline
		\end{tabular}
	\end{center}
	\vspace{-3mm}
	\caption{\small{Question-type analysis of the NDCG score achieved by different models on the val v1.0 set. } \label{tab:ques_type_analysis}}
	\vspace{-3mm}
\end{table*}

\subsection{Qualitative Analysis}
%

In addition to the examples illustrated in Figure~\ref{fig:illustration}, Figure~\ref{fig:example_visual_attention} provide six more examples to visualize the learned attention maps. The associated dialog histories are omitted for simplicity.
Typically, the attention maps become sharper and more focused throughout the reasoning process. During multiple steps, the model gradually learns to narrow down to the image regions of key objects relevant to the questions (``\emph{laptops}'', ``\emph{stove}'', ``\emph{sneakers}'', ``\emph{hat}'', ``\emph{dog's eyes}'' and ``\emph{woman's clothes}''). For instance, in the top-right example, the model focuses on the wrong region (``\emph{man}'') in the 1st step, but gradually shifts its focus to the correct regions (``\emph{dog's eyes}'') in the later steps. 





\subsection{Visual Dialog Challenge 2019}

Now, we discuss how we further boost the performance of ReDAN for participating Visual Dialog Challenge 2019\footnote{\url{https://visualdialog.org/challenge/2019}}. 

\paragraph{Rank Aggregation}
As shown in Table~\ref{tab:results_val} and~\ref{tab:results_val_gene}, ensemble of discriminative  or generative models increase the NDCG score to some extent. Empirically, we found that aggregating the ranking results of both discriminative and generative models readily boost the performance. Results are summarized in Table~\ref{tab:ens}. Combining one discriminative and one generative model already shows much better NDCG results than ensemble of 4 discriminative models. The ensemble of 4 discriminative and 4 generative models further boosts the performance. It is interesting to note that using average of the ranks results in better NDCG than using reciprocal of the ranks, though the reciprocal method achieves better results on the other metrics. Since NDCG is the metric we mostly care about, the method of averaging ranking results from different models is adopted. 

Finally, we have tried using different image feature inputs, and incorporating relation-aware encoders~\cite{li2019relation} into ReDAN to further boost the performance. By this diverse set of ensembles (called ReDAN+), we achieve an NDCG score of 67.12\% on the val v1.0 set.

\paragraph{Results on VisDial test-std v1.0}
We also evaluate the proposed ReDAN on the blind test-std v1.0 set, by submitting results to the online evaluation server. Table~\ref{tab:result_test} shows the comparison between our model and state-of-the-art visual dialog models. By using a diverse set of ensembles, ReDAN+ outperforms the state of the art method, DAN~\cite{kottur2018visual}, by a significant margin, lifting NDCG from 59.36\% to 64.47\%.

\paragraph{Question-Type Analysis}
%


%
We further perform a question-type analysis of the NDCG scores achieved by different models. We classify questions into 4 categories: \emph{Yes/no}, \emph{Number}, \emph{Color}, and \emph{Others}. 
As illustrated in Table~\ref{tab:ques_type_analysis}, in terms of the NDCG score, generative models performed better on Yes/no questions, while discriminative models performed better on all the other types of questions. We hypothesize that this is due to that generative models tend to ranking short answers higher, thus is beneficial for Yes/no questions.
Since Yes/no questions take a majority of all the questions (75\%), the better performance of generative models on the Yes/no questions translated into an overall better performance of generative models. 
Aggregating the ranking results of both discriminative and generative models results in the mutual enhancement of each other, and therefore boosting the final NDCG score by a large margin. Also, we observe that the Number questions are most difficult to answer, since training a model to count is a challenging research problem. 

\section{Conclusion}
%
We have presented Recurrent Dual Attention Network (ReDAN), a new multimodal framework for visual dialog, by incorporating image and dialog history context via a recurrently-updated query vector for multi-step reasoning. This iterative reasoning process enables model to achieve a fine-grained understanding of multimodal context, thus boosting question answering performance over state-of-the-art methods.
Experiments on the VisDial dataset validate the effectiveness of the proposed approach.

\paragraph{Acknowledgements}
We thank Yuwei Fang, Huazheng Wang and Junjie Hu
for helpful discussions. We thank anonymous reviewers for
their constructive feedbacks.

\bibliography{references}

\begin{thebibliography}{68}
\expandafter\ifx\csname natexlab\endcsname\relax\def\natexlab#1{#1}\fi

\bibitem[{Anderson et~al.(2018)Anderson, He, Buehler, Teney, Johnson, Gould,
  and Zhang}]{anderson2018bottom}
Peter Anderson, Xiaodong He, Chris Buehler, Damien Teney, Mark Johnson, Stephen
  Gould, and Lei Zhang. 2018.
\newblock Bottom-up and top-down attention for image captioning and visual
  question answering.
\newblock In \emph{CVPR}.

\bibitem[{Andreas et~al.(2016)Andreas, Rohrbach, Darrell, and
  Klein}]{andreas2016neural}
Jacob Andreas, Marcus Rohrbach, Trevor Darrell, and Dan Klein. 2016.
\newblock Neural module networks.
\newblock In \emph{CVPR}.

\bibitem[{Antol et~al.(2015)Antol, Agrawal, Lu, Mitchell, Batra,
  Lawrence~Zitnick, and Parikh}]{antol2015vqa}
Stanislaw Antol, Aishwarya Agrawal, Jiasen Lu, Margaret Mitchell, Dhruv Batra,
  C~Lawrence~Zitnick, and Devi Parikh. 2015.
\newblock Vqa: Visual question answering.
\newblock In \emph{ICCV}.

\bibitem[{Bahdanau et~al.(2015)Bahdanau, Cho, and Bengio}]{bahdanau2014neural}
Dzmitry Bahdanau, Kyunghyun Cho, and Yoshua Bengio. 2015.
\newblock Neural machine translation by jointly learning to align and
  translate.
\newblock In \emph{ICLR}.

\bibitem[{Chattopadhyay et~al.(2017)Chattopadhyay, Yadav, Prabhu,
  Chandrasekaran, Das, Lee, Batra, and Parikh}]{chattopadhyay2017evaluating}
Prithvijit Chattopadhyay, Deshraj Yadav, Viraj Prabhu, Arjun Chandrasekaran,
  Abhishek Das, Stefan Lee, Dhruv Batra, and Devi Parikh. 2017.
\newblock Evaluating visual conversational agents via cooperative human-ai
  games.
\newblock In \emph{HCOMP}.

\bibitem[{Chen et~al.(2018)Chen, Shen, Gao, Liu, and Liu}]{chen2017language}
Jianbo Chen, Yelong Shen, Jianfeng Gao, Jingjing Liu, and Xiaodong Liu. 2018.
\newblock Language-based image editing with recurrent attentive models.
\newblock In \emph{CVPR}.

\bibitem[{Cho et~al.(2014)Cho, Van~Merri{\"e}nboer, Gulcehre, Bahdanau,
  Bougares, Schwenk, and Bengio}]{cho2014learning}
Kyunghyun Cho, Bart Van~Merri{\"e}nboer, Caglar Gulcehre, Dzmitry Bahdanau,
  Fethi Bougares, Holger Schwenk, and Yoshua Bengio. 2014.
\newblock Learning phrase representations using rnn encoder-decoder for
  statistical machine translation.
\newblock \emph{arXiv preprint arXiv:1406.1078}.

\bibitem[{Choi et~al.(2018)Choi, He, Iyyer, Yatskar, Yih, Choi, Liang, and
  Zettlemoyer}]{choi2018quac}
Eunsol Choi, He~He, Mohit Iyyer, Mark Yatskar, Wen-tau Yih, Yejin Choi, Percy
  Liang, and Luke Zettlemoyer. 2018.
\newblock Quac: Question answering in context.
\newblock In \emph{EMNLP}.

\bibitem[{Cui et~al.(2017)Cui, Chen, Wei, Wang, Liu, and Hu}]{cui2016attention}
Yiming Cui, Zhipeng Chen, Si~Wei, Shijin Wang, Ting Liu, and Guoping Hu. 2017.
\newblock Attention-over-attention neural networks for reading comprehension.
\newblock In \emph{ACL}.

\bibitem[{Das et~al.(2017{\natexlab{a}})Das, Kottur, Gupta, Singh, Yadav,
  Moura, Parikh, and Batra}]{das2017visual}
Abhishek Das, Satwik Kottur, Khushi Gupta, Avi Singh, Deshraj Yadav,
  Jos{\'e}~MF Moura, Devi Parikh, and Dhruv Batra. 2017{\natexlab{a}}.
\newblock Visual dialog.
\newblock In \emph{CVPR}.

\bibitem[{Das et~al.(2017{\natexlab{b}})Das, Kottur, Moura, Lee, and
  Batra}]{das2017learning}
Abhishek Das, Satwik Kottur, Jos{\'e}~MF Moura, Stefan Lee, and Dhruv Batra.
  2017{\natexlab{b}}.
\newblock Learning cooperative visual dialog agents with deep reinforcement
  learning.
\newblock In \emph{ICCV}.

\bibitem[{De~Vries et~al.(2017)De~Vries, Strub, Chandar, Pietquin, Larochelle,
  and Courville}]{de2017guesswhat}
Harm De~Vries, Florian Strub, Sarath Chandar, Olivier Pietquin, Hugo
  Larochelle, and Aaron~C Courville. 2017.
\newblock Guesswhat?! visual object discovery through multi-modal dialogue.
\newblock In \emph{CVPR}.

\bibitem[{Dhingra et~al.(2017)Dhingra, Liu, Yang, Cohen, and
  Salakhutdinov}]{dhingra2016gated}
Bhuwan Dhingra, Hanxiao Liu, Zhilin Yang, William~W Cohen, and Ruslan
  Salakhutdinov. 2017.
\newblock Gated-attention readers for text comprehension.
\newblock In \emph{ACL}.

\bibitem[{Fang et~al.(2015)Fang, Gupta, Iandola, Srivastava, Deng, Doll{\'a}r,
  Gao, He, Mitchell, Platt et~al.}]{fang2015captions}
Hao Fang, Saurabh Gupta, Forrest Iandola, Rupesh~K Srivastava, Li~Deng, Piotr
  Doll{\'a}r, Jianfeng Gao, Xiaodong He, Margaret Mitchell, John~C Platt,
  et~al. 2015.
\newblock From captions to visual concepts and back.
\newblock In \emph{CVPR}.

\bibitem[{Fukui et~al.(2016)Fukui, Park, Yang, Rohrbach, Darrell, and
  Rohrbach}]{fukui2016multimodal}
Akira Fukui, Dong~Huk Park, Daylen Yang, Anna Rohrbach, Trevor Darrell, and
  Marcus Rohrbach. 2016.
\newblock Multimodal compact bilinear pooling for visual question answering and
  visual grounding.
\newblock In \emph{EMNLP}.

\bibitem[{Gao et~al.(2018)Gao, Galley, and Li}]{gaosurvey}
Jianfeng Gao, Michel Galley, and Lihong Li. 2018.
\newblock Neural approaches to conversational ai.
\newblock \emph{arXiv preprint arXiv:1809.08267}.

\bibitem[{Goodfellow et~al.(2014)Goodfellow, Pouget-Abadie, Mirza, Xu,
  Warde-Farley, Ozair, Courville, and Bengio}]{goodfellow2014generative}
Ian Goodfellow, Jean Pouget-Abadie, Mehdi Mirza, Bing Xu, David Warde-Farley,
  Sherjil Ozair, Aaron Courville, and Yoshua Bengio. 2014.
\newblock Generative adversarial nets.
\newblock In \emph{NIPS}.

\bibitem[{Gregor et~al.(2015)Gregor, Danihelka, Graves, Rezende, and
  Wierstra}]{gregor2015draw}
Karol Gregor, Ivo Danihelka, Alex Graves, Danilo~Jimenez Rezende, and Daan
  Wierstra. 2015.
\newblock Draw: A recurrent neural network for image generation.
\newblock In \emph{ICML}.

\bibitem[{Guo et~al.(2019)Guo, Xu, and Tao}]{guo2019image}
Dalu Guo, Chang Xu, and Dacheng Tao. 2019.
\newblock Image-question-answer synergistic network for visual dialog.
\newblock \emph{arXiv preprint arXiv:1902.09774}.

\bibitem[{Guo et~al.(2018)Guo, Wu, Cheng, Rennie, and Feris}]{guo2018dialog}
Xiaoxiao Guo, Hui Wu, Yu~Cheng, Steven Rennie, and Rogerio~Schmidt Feris. 2018.
\newblock Dialog-based interactive image retrieval.
\newblock In \emph{NIPS}.

\bibitem[{He et~al.(2016)He, Zhang, Ren, and Sun}]{he2016deep}
Kaiming He, Xiangyu Zhang, Shaoqing Ren, and Jian Sun. 2016.
\newblock Deep residual learning for image recognition.
\newblock In \emph{CVPR}.

\bibitem[{Hill et~al.(2016)Hill, Bordes, Chopra, and
  Weston}]{hill2015goldilocks}
Felix Hill, Antoine Bordes, Sumit Chopra, and Jason Weston. 2016.
\newblock The goldilocks principle: Reading children's books with explicit
  memory representations.
\newblock In \emph{ICLR}.

\bibitem[{Hochreiter and Schmidhuber(1997)}]{hochreiter1997long}
Sepp Hochreiter and J{\"u}rgen Schmidhuber. 1997.
\newblock Long short-term memory.
\newblock \emph{Neural computation}.

\bibitem[{Hudson and Manning(2018)}]{hudson2018compositional}
Drew~A Hudson and Christopher~D Manning. 2018.
\newblock Compositional attention networks for machine reasoning.
\newblock In \emph{ICLR}.

\bibitem[{Jain et~al.(2018)Jain, Lazebnik, and Schwing}]{jain2018two}
Unnat Jain, Svetlana Lazebnik, and Alexander~G Schwing. 2018.
\newblock Two can play this game: visual dialog with discriminative question
  generation and answering.
\newblock In \emph{CVPR}.

\bibitem[{Kang et~al.(2019)Kang, Lim, and Zhang}]{kang2019dual}
Gi-Cheon Kang, Jaeseo Lim, and Byoung-Tak Zhang. 2019.
\newblock Dual attention networks for visual reference resolution in visual
  dialog.
\newblock \emph{arXiv preprint arXiv:1902.09368}.

\bibitem[{Kim et~al.(2017)Kim, On, Lim, Kim, Ha, and Zhang}]{kim2016hadamard}
Jin-Hwa Kim, Kyoung-Woon On, Woosang Lim, Jeonghee Kim, Jung-Woo Ha, and
  Byoung-Tak Zhang. 2017.
\newblock Hadamard product for low-rank bilinear pooling.
\newblock In \emph{ICLR}.

\bibitem[{Kingma and Ba(2014)}]{kingma2014adam}
Diederik~P Kingma and Jimmy Ba. 2014.
\newblock Adam: A method for stochastic optimization.
\newblock \emph{arXiv preprint arXiv:1412.6980}.

\bibitem[{Kingma and Welling(2014)}]{kingma2013auto}
Diederik~P Kingma and Max Welling. 2014.
\newblock Auto-encoding variational bayes.
\newblock In \emph{ICLR}.

\bibitem[{Kottur et~al.(2018)Kottur, Moura, Parikh, Batra, and
  Rohrbach}]{kottur2018visual}
Satwik Kottur, Jos{\'e}~MF Moura, Devi Parikh, Dhruv Batra, and Marcus
  Rohrbach. 2018.
\newblock Visual coreference resolution in visual dialog using neural module
  networks.
\newblock In \emph{ECCV}.

\bibitem[{Krishna et~al.(2017)Krishna, Zhu, Groth, Johnson, Hata, Kravitz,
  Chen, Kalantidis, Li, Shamma et~al.}]{krishna2017visual}
Ranjay Krishna, Yuke Zhu, Oliver Groth, Justin Johnson, Kenji Hata, Joshua
  Kravitz, Stephanie Chen, Yannis Kalantidis, Li-Jia Li, David~A Shamma, et~al.
  2017.
\newblock Visual genome: Connecting language and vision using crowdsourced
  dense image annotations.
\newblock \emph{IJCV}.

\bibitem[{Lee et~al.(2018)Lee, Heo, and Zhang}]{lee2018answerer}
Sang-Woo Lee, Yu-Jung Heo, and Byoung-Tak Zhang. 2018.
\newblock Answerer in questioner's mind for goal-oriented visual dialogue.
\newblock In \emph{NIPS}.

\bibitem[{Li et~al.(2017)Li, Monroe, Shi, Jean, Ritter, and
  Jurafsky}]{li2017adversarial}
Jiwei Li, Will Monroe, Tianlin Shi, S{\'e}bastien Jean, Alan Ritter, and Dan
  Jurafsky. 2017.
\newblock Adversarial learning for neural dialogue generation.
\newblock In \emph{EMNLP}.

\bibitem[{Li et~al.(2019)Li, Gan, Cheng, and Liu}]{li2019relation}
Linjie Li, Zhe Gan, Yu~Cheng, and Jingjing Liu. 2019.
\newblock Relation-aware graph attention network for visual question answering.
\newblock \emph{arXiv preprint arXiv:1903.12314}.

\bibitem[{Lin et~al.(2014)Lin, Maire, Belongie, Hays, Perona, Ramanan,
  Doll{\'a}r, and Zitnick}]{lin2014microsoft}
Tsung-Yi Lin, Michael Maire, Serge Belongie, James Hays, Pietro Perona, Deva
  Ramanan, Piotr Doll{\'a}r, and C~Lawrence Zitnick. 2014.
\newblock Microsoft coco: Common objects in context.
\newblock In \emph{ECCV}.

\bibitem[{Liu et~al.(2018)Liu, Shen, Duh, and Gao}]{liu2017stochastic}
Xiaodong Liu, Yelong Shen, Kevin Duh, and Jianfeng Gao. 2018.
\newblock Stochastic answer networks for machine reading comprehension.
\newblock In \emph{ACL}.

\bibitem[{Lu et~al.(2017)Lu, Kannan, Yang, Parikh, and Batra}]{lu2017best}
Jiasen Lu, Anitha Kannan, Jianwei Yang, Devi Parikh, and Dhruv Batra. 2017.
\newblock Best of both worlds: Transferring knowledge from discriminative
  learning to a generative visual dialog model.
\newblock In \emph{NIPS}.

\bibitem[{Massiceti et~al.(2018)Massiceti, Siddharth, Dokania, and
  Torr}]{massiceti2018flipdial}
Daniela Massiceti, N~Siddharth, Puneet~K Dokania, and Philip~HS Torr. 2018.
\newblock Flipdial: A generative model for two-way visual dialogue.
\newblock In \emph{CVPR}.

\bibitem[{Mnih et~al.(2014)Mnih, Heess, Graves et~al.}]{mnih2014recurrent}
Volodymyr Mnih, Nicolas Heess, Alex Graves, et~al. 2014.
\newblock Recurrent models of visual attention.
\newblock In \emph{NIPS}.

\bibitem[{Mostafazadeh et~al.(2017)Mostafazadeh, Brockett, Dolan, Galley, Gao,
  Spithourakis, and Vanderwende}]{mostafazadeh2017image}
Nasrin Mostafazadeh, Chris Brockett, Bill Dolan, Michel Galley, Jianfeng Gao,
  Georgios~P Spithourakis, and Lucy Vanderwende. 2017.
\newblock Image-grounded conversations: Multimodal context for natural question
  and response generation.
\newblock \emph{arXiv preprint arXiv:1701.08251}.

\bibitem[{Nam et~al.(2017)Nam, Ha, and Kim}]{nam2016dual}
Hyeonseob Nam, Jung-Woo Ha, and Jeonghee Kim. 2017.
\newblock Dual attention networks for multimodal reasoning and matching.
\newblock In \emph{CVPR}.

\bibitem[{Niu et~al.(2018)Niu, Zhang, Zhang, Zhang, Lu, and
  Wen}]{niu2018recursive}
Yulei Niu, Hanwang Zhang, Manli Zhang, Jianhong Zhang, Zhiwu Lu, and Ji-Rong
  Wen. 2018.
\newblock Recursive visual attention in visual dialog.
\newblock \emph{arXiv preprint arXiv:1812.02664}.

\bibitem[{Pennington et~al.(2014)Pennington, Socher, and
  Manning}]{pennington2014glove}
Jeffrey Pennington, Richard Socher, and Christopher Manning. 2014.
\newblock Glove: Global vectors for word representation.
\newblock In \emph{EMNLP}.

\bibitem[{Reddy et~al.(2018)Reddy, Chen, and Manning}]{reddy2018coqa}
Siva Reddy, Danqi Chen, and Christopher~D Manning. 2018.
\newblock Coqa: A conversational question answering challenge.
\newblock In \emph{EMNLP}.

\bibitem[{Ren et~al.(2015)Ren, He, Girshick, and Sun}]{ren2015faster}
Shaoqing Ren, Kaiming He, Ross Girshick, and Jian Sun. 2015.
\newblock Faster r-cnn: Towards real-time object detection with region proposal
  networks.
\newblock In \emph{NIPS}.

\bibitem[{Schwartz et~al.(2019)Schwartz, Yu, Hazan, and
  Schwing}]{schwartz2019factor}
Idan Schwartz, Seunghak Yu, Tamir Hazan, and Alexander Schwing. 2019.
\newblock Factor graph attention.
\newblock \emph{arXiv preprint arXiv:1904.05880}.

\bibitem[{Seo et~al.(2017)Seo, Lehrmann, Han, and Sigal}]{seo2017visual}
Paul~Hongsuck Seo, Andreas Lehrmann, Bohyung Han, and Leonid Sigal. 2017.
\newblock Visual reference resolution using attention memory for visual dialog.
\newblock In \emph{NIPS}.

\bibitem[{Shekhar et~al.(2018)Shekhar, Baumgartner, Venkatesh, Bruni, Bernardi,
  and Fernandez}]{shekhar2018ask}
Ravi Shekhar, Tim Baumgartner, Aashish Venkatesh, Elia Bruni, Raffaella
  Bernardi, and Raquel Fernandez. 2018.
\newblock Ask no more: Deciding when to guess in referential visual dialogue.
\newblock In \emph{COLING}.

\bibitem[{Shen et~al.(2017)Shen, Huang, Gao, and Chen}]{shen2017reasonet}
Yelong Shen, Po-Sen Huang, Jianfeng Gao, and Weizhu Chen. 2017.
\newblock Reasonet: Learning to stop reading in machine comprehension.
\newblock In \emph{KDD}.

\bibitem[{Simonyan and Zisserman(2014)}]{simonyan2014very}
Karen Simonyan and Andrew Zisserman. 2014.
\newblock Very deep convolutional networks for large-scale image recognition.
\newblock \emph{arXiv preprint arXiv:1409.1556}.

\bibitem[{Sohn et~al.(2015)Sohn, Lee, and Yan}]{sohn2015learning}
Kihyuk Sohn, Honglak Lee, and Xinchen Yan. 2015.
\newblock Learning structured output representation using deep conditional
  generative models.
\newblock In \emph{NIPS}.

\bibitem[{Sordoni et~al.(2016)Sordoni, Bachman, Trischler, and
  Bengio}]{sordoni2016iterative}
Alessandro Sordoni, Philip Bachman, Adam Trischler, and Yoshua Bengio. 2016.
\newblock Iterative alternating neural attention for machine reading.
\newblock \emph{arXiv preprint arXiv:1606.02245}.

\bibitem[{Srivastava et~al.(2014)Srivastava, Hinton, Krizhevsky, Sutskever, and
  Salakhutdinov}]{srivastava2014dropout}
Nitish Srivastava, Geoffrey Hinton, Alex Krizhevsky, Ilya Sutskever, and Ruslan
  Salakhutdinov. 2014.
\newblock Dropout: a simple way to prevent neural networks from overfitting.
\newblock \emph{JMLR}.

\bibitem[{Strub et~al.(2017)Strub, De~Vries, Mary, Piot, Courville, and
  Pietquin}]{strub2017end}
Florian Strub, Harm De~Vries, Jeremie Mary, Bilal Piot, Aaron Courville, and
  Olivier Pietquin. 2017.
\newblock End-to-end optimization of goal-driven and visually grounded dialogue
  systems.
\newblock In \emph{IJCAI}.

\bibitem[{Strub et~al.(2018)Strub, Seurin, Perez, de~Vries, Preux, Courville,
  Pietquin et~al.}]{strub2018visual}
Florian Strub, Mathieu Seurin, Ethan Perez, Harm de~Vries, Philippe Preux,
  Aaron Courville, Olivier Pietquin, et~al. 2018.
\newblock Visual reasoning with multi-hop feature modulation.
\newblock In \emph{ECCV}.

\bibitem[{Sutskever et~al.(2014)Sutskever, Vinyals, and
  Le}]{sutskever2014sequence}
Ilya Sutskever, Oriol Vinyals, and Quoc~V Le. 2014.
\newblock Sequence to sequence learning with neural networks.
\newblock In \emph{NIPS}.

\bibitem[{Teney et~al.(2018)Teney, Anderson, He, and van~den
  Hengel}]{teney2017tips}
Damien Teney, Peter Anderson, Xiaodong He, and Anton van~den Hengel. 2018.
\newblock Tips and tricks for visual question answering: Learnings from the
  2017 challenge.
\newblock In \emph{CVPR}.

\bibitem[{Vinyals et~al.(2015)Vinyals, Toshev, Bengio, and
  Erhan}]{vinyals2015show}
Oriol Vinyals, Alexander Toshev, Samy Bengio, and Dumitru Erhan. 2015.
\newblock Show and tell: A neural image caption generator.
\newblock In \emph{CVPR}.

\bibitem[{Wu et~al.(2018)Wu, Wang, Shen, Reid, and van~den Hengel}]{wu2017you}
Qi~Wu, Peng Wang, Chunhua Shen, Ian Reid, and Anton van~den Hengel. 2018.
\newblock Are you talking to me? reasoned visual dialog generation through
  adversarial learning.
\newblock In \emph{CVPR}.

\bibitem[{Xu et~al.(2015)Xu, Ba, Kiros, Cho, Courville, Salakhudinov, Zemel,
  and Bengio}]{xu2015show}
Kelvin Xu, Jimmy Ba, Ryan Kiros, Kyunghyun Cho, Aaron Courville, Ruslan
  Salakhudinov, Rich Zemel, and Yoshua Bengio. 2015.
\newblock Show, attend and tell: Neural image caption generation with visual
  attention.
\newblock In \emph{ICML}.

\bibitem[{Yang et~al.(2019)Yang, Zha, and Zhang}]{yang2019making}
Tianhao Yang, Zheng-Jun Zha, and Hanwang Zhang. 2019.
\newblock Making history matter: Gold-critic sequence training for visual
  dialog.
\newblock \emph{arXiv preprint arXiv:1902.09326}.

\bibitem[{Yang et~al.(2016)Yang, He, Gao, Deng, and Smola}]{yang2016stacked}
Zichao Yang, Xiaodong He, Jianfeng Gao, Li~Deng, and Alex Smola. 2016.
\newblock Stacked attention networks for image question answering.
\newblock In \emph{CVPR}.

\bibitem[{Yu et~al.(2017{\natexlab{a}})Yu, Lee, and Le}]{yu2017learning}
Adams~Wei Yu, Hongrae Lee, and Quoc~V Le. 2017{\natexlab{a}}.
\newblock Learning to skim text.
\newblock \emph{arXiv preprint arXiv:1704.06877}.

\bibitem[{Yu et~al.(2017{\natexlab{b}})Yu, Zhang, Wang, and Yu}]{yu2017seqgan}
Lantao Yu, Weinan Zhang, Jun Wang, and Yong Yu. 2017{\natexlab{b}}.
\newblock Seqgan: Sequence generative adversarial nets with policy gradient.
\newblock In \emph{AAAI}.

\bibitem[{Yu et~al.(2017{\natexlab{c}})Yu, Yu, Fan, and Tao}]{yu2017multi}
Zhou Yu, Jun Yu, Jianping Fan, and Dacheng Tao. 2017{\natexlab{c}}.
\newblock Multi-modal factorized bilinear pooling with co-attention learning
  for visual question answering.
\newblock In \emph{ICCV}.

\bibitem[{Zhang et~al.(2019)Zhang, Ghosh, Heck, Walsh, Zhang, Zhang, and
  Kuo}]{zhang2019generative}
Heming Zhang, Shalini Ghosh, Larry Heck, Stephen Walsh, Junting Zhang, Jie
  Zhang, and C-C~Jay Kuo. 2019.
\newblock Generative visual dialogue system via adaptive reasoning and weighted
  likelihood estimation.
\newblock \emph{arXiv preprint arXiv:1902.09818}.

\bibitem[{Zhang et~al.(2018)Zhang, Wu, Shen, Zhang, Lu, and Van
  Den~Hengel}]{zhang2018goal}
Junjie Zhang, Qi~Wu, Chunhua Shen, Jian Zhang, Jianfeng Lu, and Anton Van
  Den~Hengel. 2018.
\newblock Goal-oriented visual question generation via intermediate rewards.
\newblock In \emph{ECCV}.

\bibitem[{Zheng et~al.(2019)Zheng, Wang, Qi, and Zhu}]{zheng2019reasoning}
Zilong Zheng, Wenguan Wang, Siyuan Qi, and Song-Chun Zhu. 2019.
\newblock Reasoning visual dialogs with structural and partial observations.
\newblock \emph{arXiv preprint arXiv:1904.05548}.

\end{thebibliography}
\bibliographystyle{acl_natbib}

\newpage
\hspace{2mm}
\newpage

\appendix

\begin{figure*}
	\centering
	\includegraphics[width=1.00\textwidth]{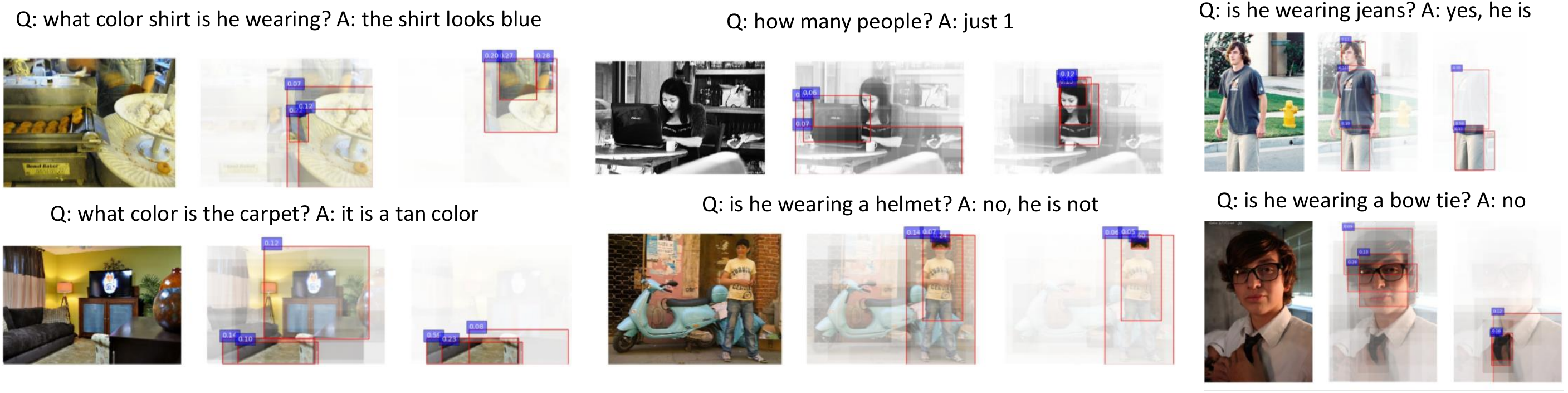} \\
	\includegraphics[width=1.00\textwidth]{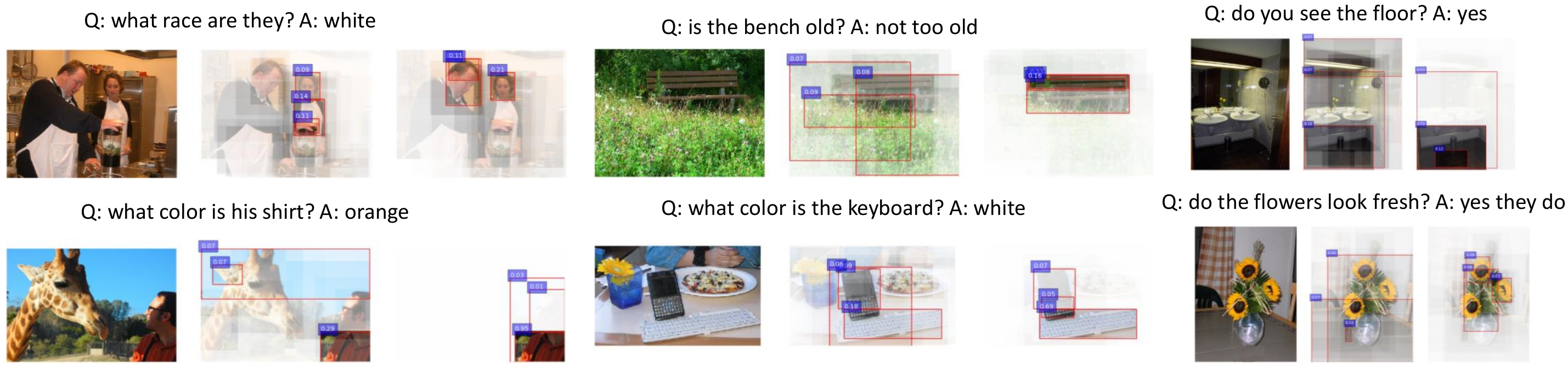} \\
	\includegraphics[width=1.00\textwidth]{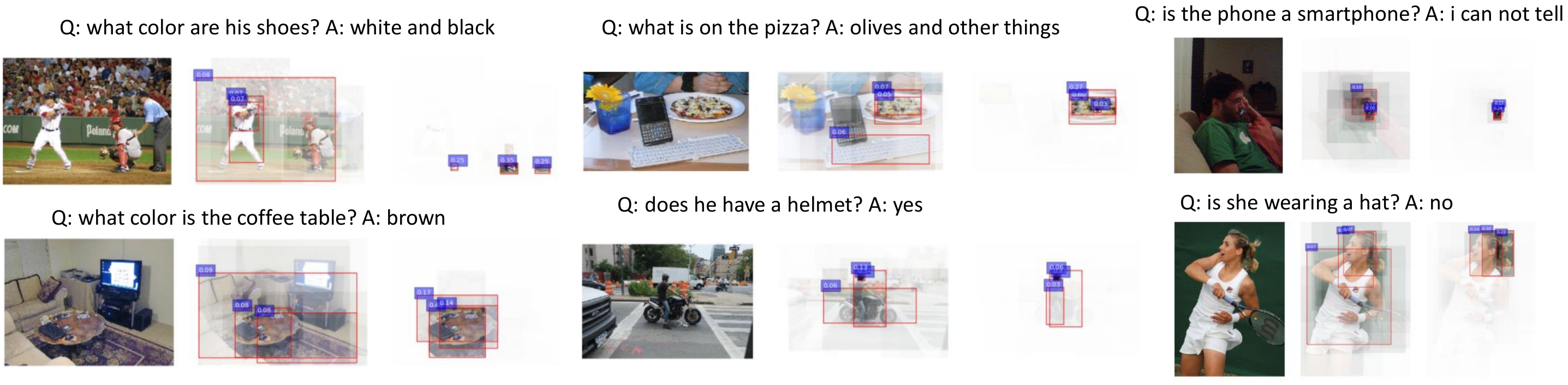} \\
	\includegraphics[width=1.00\textwidth]{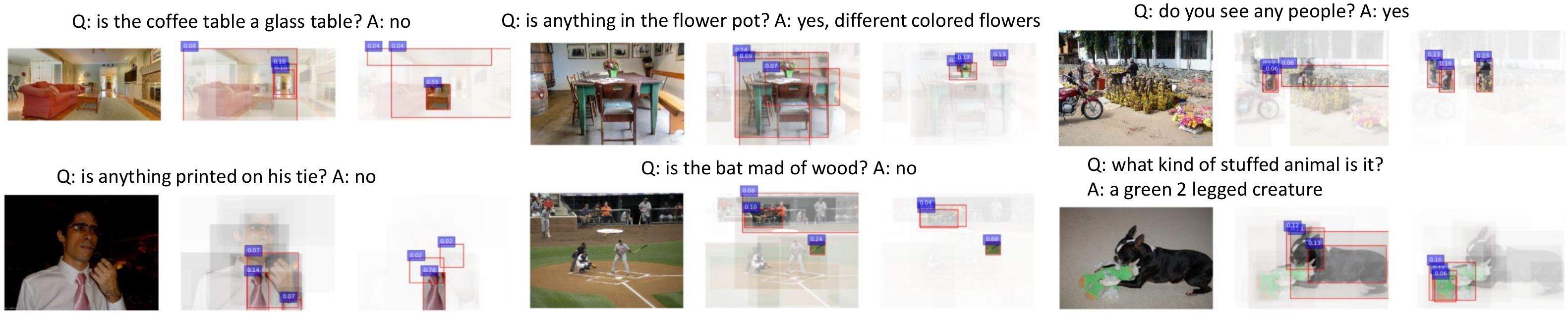}
	\caption{\small{Visualization of learned attention maps using 2 reasoning steps.}}
\end{figure*}

\begin{figure*}[h]
	\centering
	\includegraphics[width=1.00\textwidth]{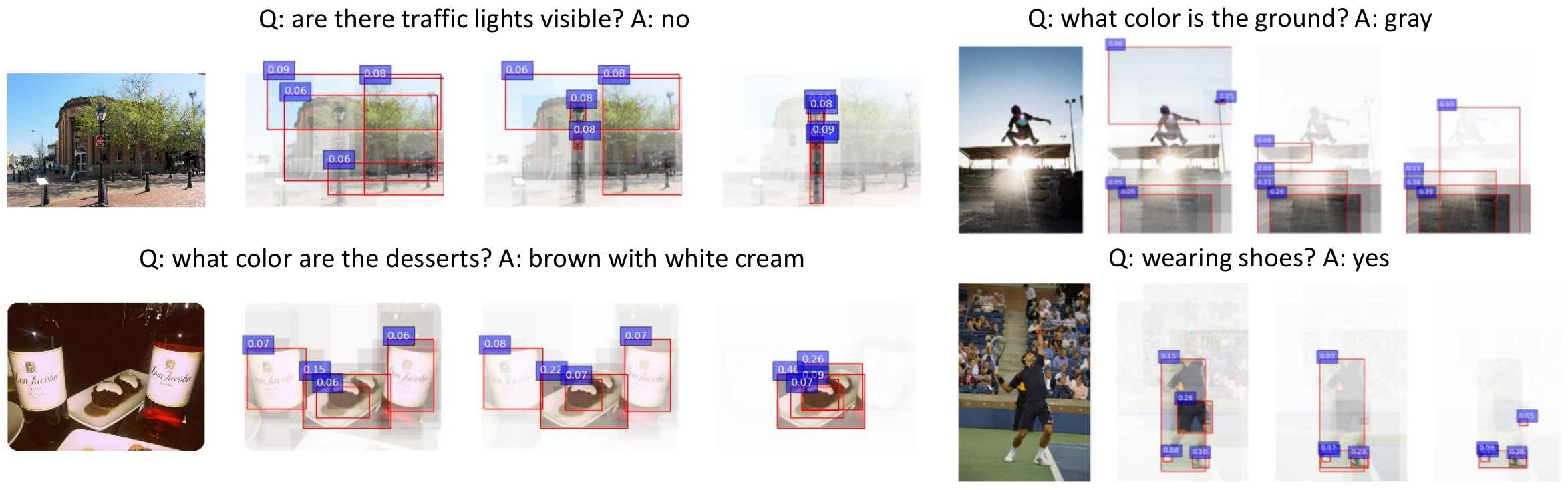} \\
	\includegraphics[width=1.00\textwidth]{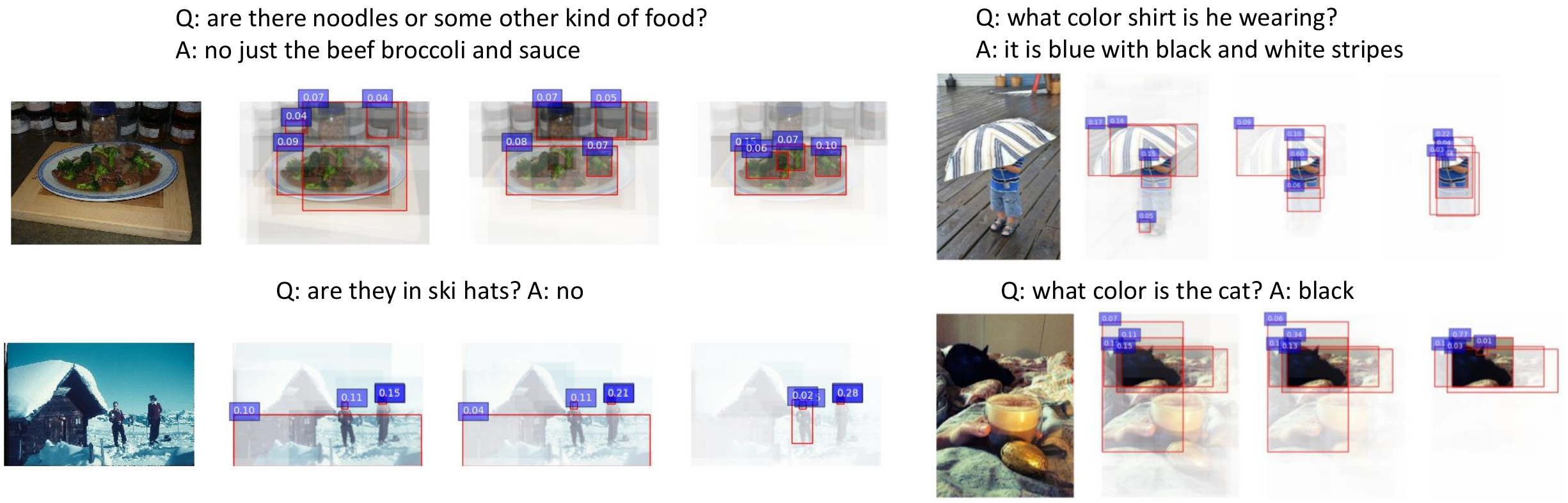} \\
	\includegraphics[width=1.00\textwidth]{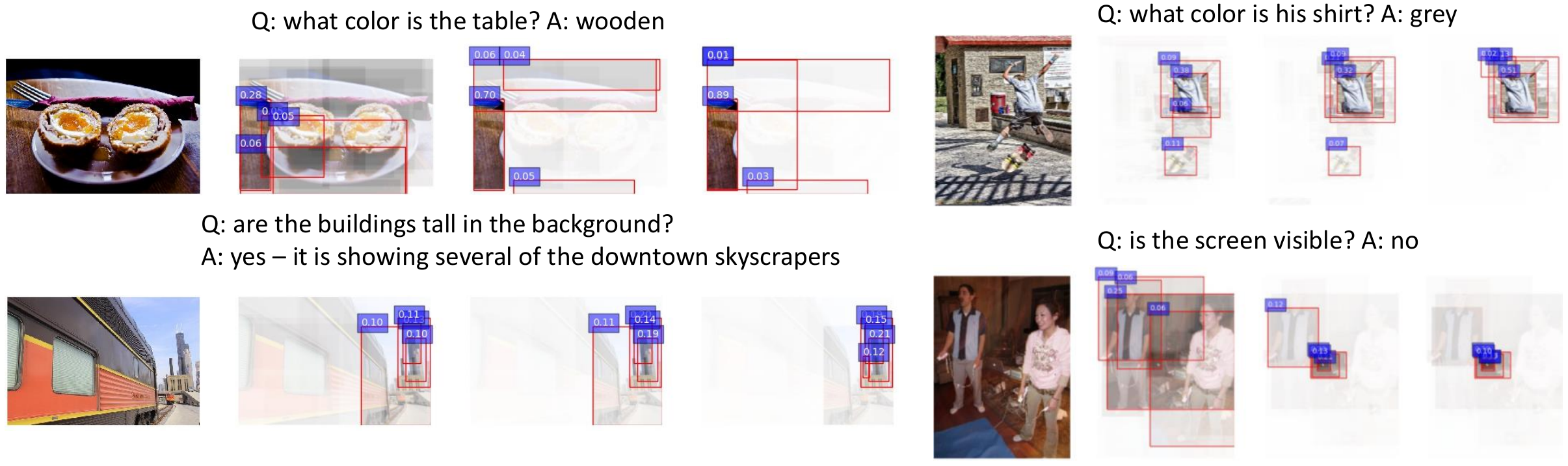}
	\caption{\small{Visualization of learned attention maps using 3 reasoning steps.}}
	\label{fig:example_img_cap}
	\vspace{-2mm}
\end{figure*}



\end{document}